\documentclass[letterpaper, 10 pt, conference]{ieeeconf}  

\IEEEoverridecommandlockouts                              

\renewcommand{\textfloatsep}{0em}

\usepackage{graphics} 
\usepackage{epsfig} 

\usepackage{times} 
\usepackage{amsmath} 
\usepackage{amssymb}  

\usepackage{subcaption}
\usepackage{color}
\usepackage{hyperref}
\usepackage{framed}
\usepackage{listings}

\def\secref#1{Sec.~\ref{#1}}
\def\figref#1{Fig.~\ref{#1}}
\def\tabref#1{Tab.~\ref{#1}}
\def\eqref#1{Eq.~(\ref{#1})}

\def\lstref#1{Lst.~\ref{#1}}
\def\ourmethod{ProSLAM}

\definecolor{shadecolor}{rgb}{0.95, 0.95, 0.95}
\definecolor{orange}{rgb}{1.0, 0.5, 0.0}
\definecolor{indigo}{rgb}{0.5, 0.0, 0.5}

  \makeatletter
  \providecommand\color[2][]{%
    \GenericError{(gnuplot) \space\space\space\@spaces}{%
      Package color not loaded in conjunction with
      terminal option `colourtext'%
    }{See the gnuplot documentation for explanation.%
    }{Either use 'blacktext' in gnuplot or load the package
      color.sty in LaTeX.}%
    \renewcommand\color[2][]{}%
  }%
  \providecommand\includegraphics[2][]{%
    \GenericError{(gnuplot) \space\space\space\@spaces}{%
      Package graphicx or graphics not loaded%
    }{See the gnuplot documentation for explanation.%
    }{The gnuplot epslatex terminal needs graphicx.sty or graphics.sty.}%
    \renewcommand\includegraphics[2][]{}%
  }%
  \providecommand\rotatebox[2]{#2}%
  \@ifundefined{ifGPcolor}{%
    \newif\ifGPcolor
    \GPcolortrue
  }{}%
  \@ifundefined{ifGPblacktext}{%
    \newif\ifGPblacktext
    \GPblacktexttrue
  }{}%
  \let\gplgaddtomacro\g@addto@macro
  \gdef\gplbacktext{}%
  \gdef\gplfronttext{}%
  \makeatother
  \ifGPblacktext
    \def\colorrgb#1{}%
    \def\colorgray#1{}%
  \else
    \ifGPcolor
      \def\colorrgb#1{\color[rgb]{#1}}%
      \def\colorgray#1{\color[gray]{#1}}%
      \expandafter\def\csname LTw\endcsname{\color{white}}%
      \expandafter\def\csname LTb\endcsname{\color{black}}%
      \expandafter\def\csname LTa\endcsname{\color{black}}%
      \expandafter\def\csname LT0\endcsname{\color[rgb]{1,0,0}}%
      \expandafter\def\csname LT1\endcsname{\color[rgb]{0,1,0}}%
      \expandafter\def\csname LT2\endcsname{\color[rgb]{0,0,1}}%
      \expandafter\def\csname LT3\endcsname{\color[rgb]{1,0,1}}%
      \expandafter\def\csname LT4\endcsname{\color[rgb]{0,1,1}}%
      \expandafter\def\csname LT5\endcsname{\color[rgb]{1,1,0}}%
      \expandafter\def\csname LT6\endcsname{\color[rgb]{0,0,0}}%
      \expandafter\def\csname LT7\endcsname{\color[rgb]{1,0.3,0}}%
      \expandafter\def\csname LT8\endcsname{\color[rgb]{0.5,0.5,0.5}}%
    \else
      \def\colorrgb#1{\color{black}}%
      \def\colorgray#1{\color[gray]{#1}}%
      \expandafter\def\csname LTw\endcsname{\color{white}}%
      \expandafter\def\csname LTb\endcsname{\color{black}}%
      \expandafter\def\csname LTa\endcsname{\color{black}}%
      \expandafter\def\csname LT0\endcsname{\color{black}}%
      \expandafter\def\csname LT1\endcsname{\color{black}}%
      \expandafter\def\csname LT2\endcsname{\color{black}}%
      \expandafter\def\csname LT3\endcsname{\color{black}}%
      \expandafter\def\csname LT4\endcsname{\color{black}}%
      \expandafter\def\csname LT5\endcsname{\color{black}}%
      \expandafter\def\csname LT6\endcsname{\color{black}}%
      \expandafter\def\csname LT7\endcsname{\color{black}}%
      \expandafter\def\csname LT8\endcsname{\color{black}}%
    \fi
  \fi
    \ifx\gptboxheight\undefined%
      \newlength{\gptboxheight}%
      \newsavebox{\gptboxtext}%
    \fi%
\title{\LARGE \bf \ourmethod: Graph SLAM from a Programmer's Perspective}
\author{Dominik Schlegel, Mirco Colosi and Giorgio Grisetti
\thanks{All authors are with the Dept. of Computer, Control and Management Engineering,
        Sapienza University of Rome, Rome, Italy {\tt\small schlegel@diag.uniroma1.it}
        {\tt\small colosi.1650227@studenti.uniroma1.it}
        {\tt\small grisetti@diag.uniroma1.it}}%
}

\begin{document}

\maketitle
\thispagestyle{empty}
\pagestyle{empty}

\begin{abstract}
  In this paper we present \ourmethod, a lightweight stereo visual
  SLAM system designed with simplicity in mind. Our work stems from
  the experience gathered by the authors while teaching SLAM to
  students and aims at providing a highly modular system
  that can be easily implemented and understood. Rather than focusing
  on the well known mathematical aspects of Stereo Visual SLAM, in
  this work we highlight the data structures and the
  algorithmic aspects that one needs to tackle during the design of
  such a system. We implemented \ourmethod~using the C++ programming
  language in combination with a minimal set of well known used
  external libraries. In addition to an open source implementation\footnote{\url{https://gitlab.com/srrg-software/srrg_proslam}},
  we provide several code snippets that address the
  core aspects of our approach directly in this paper. The results of a thorough
  validation performed on standard benchmark datasets show that our
  approach achieves accuracy comparable to state of the art
  methods, while requiring substantially less computational resources.
\end{abstract}

\section{Introduction}
\label{sec:introduction}


Simultaneous Localization and Mapping (SLAM) systems manage to deliver
incredible results and after years of extensive investigation, this
topic still captures the imagination of many young students and
prospective researchers. Among others ORB-SLAM2~\cite{mur2016orb} and
LSD-SLAM~\cite{engel2015large} are the two main approaches that are regarded
as the state of the art in the robotic community.
The increasing sophistication of these systems generally comes at a
price of higher complexity. This fact renders those systems hard to
understand and extend for people new to the field.

In this paper we present \ourmethod~(Programmers SLAM), a complete stereo
visual SLAM system that combines well known techniques, encapsulating
them in separated components with clear interfaces.
We further provide code snippets that can be copied and pasted to realize core
functionalities of the system. \ourmethod~is implemented in C++ and
makes minimal use of the basic functionalities of well known libraries
such as Eigen, for matrix calculation, OpenCV for input output
operation and feature extraction, and g2o~\cite{kummerle2011g} for
pose graph optimization. 

\begin{figure}[t]
    \centering
  \begin{subfigure}[t]{0.99\columnwidth}
    \includegraphics[width=0.99\columnwidth]{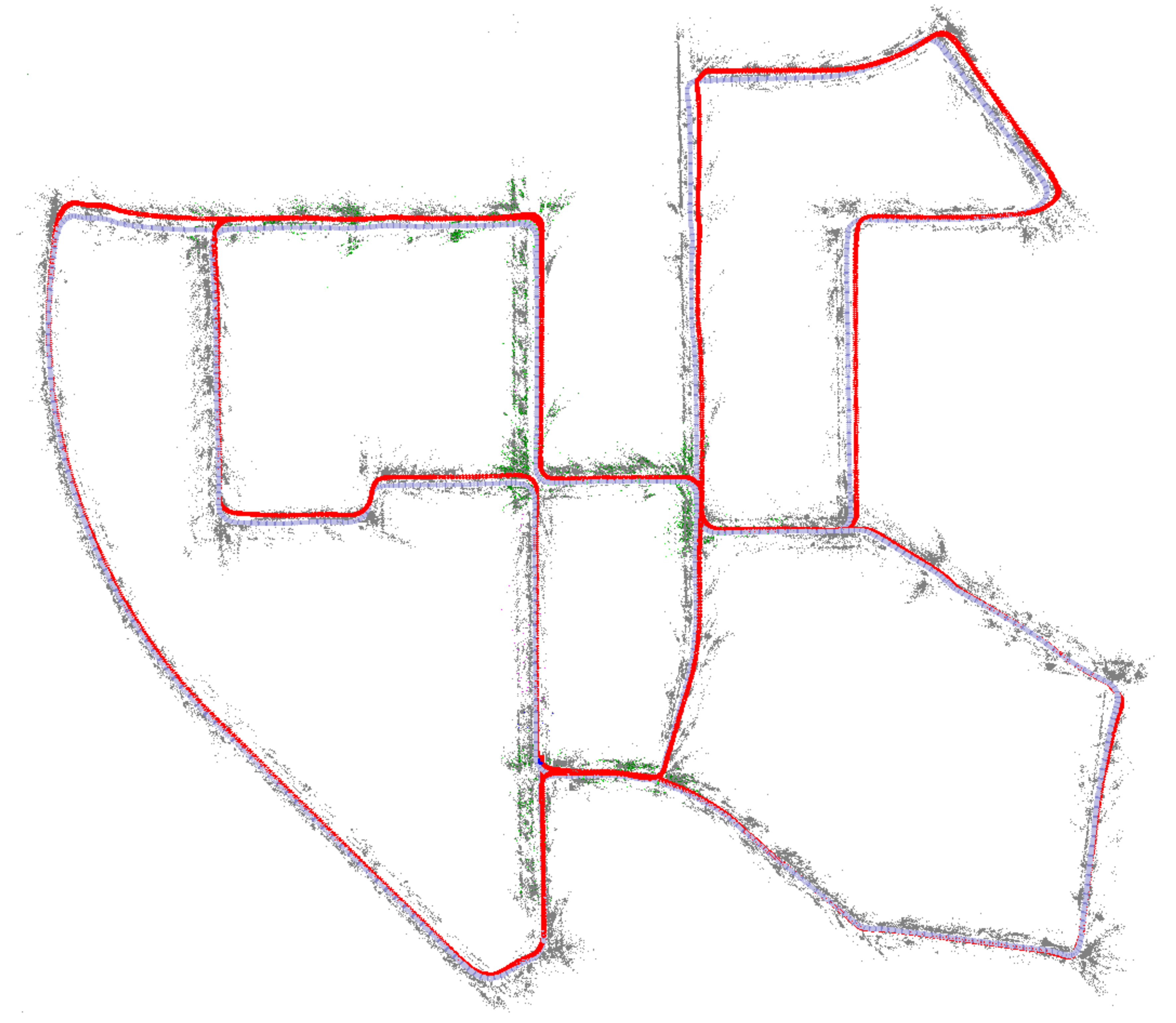}
    \caption{KITTI Sequence 00, Blue trajectory: ours, Red: ground truth}
    \label{fig:map_kitti}
  \end{subfigure}	
  \vspace{0pt}
    \begin{subfigure}[t]{0.99\columnwidth}
    \includegraphics[width=0.99\columnwidth]{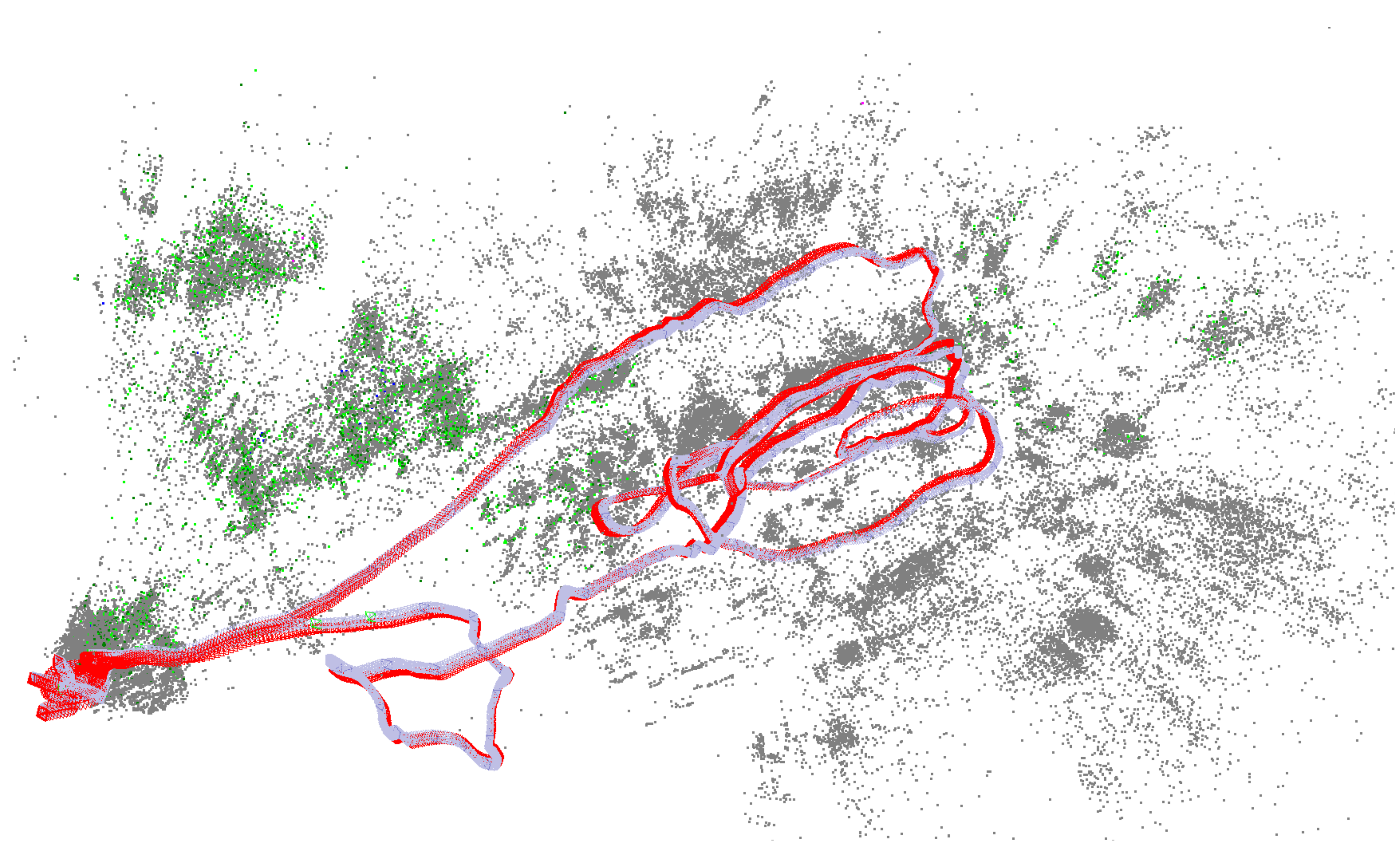}
    \caption{EuRoC MH\_01\_easy, Blue trajectory: ours, Red: ground truth}
    \label{fig:map_euroc}
  \end{subfigure}
    \caption{Final map output of \ourmethod~for two different datasets using identical parameters.
    The robot trajectory is illustrated by the light blue rectangular frame trail.
    Loop closed point clouds are highlighted in green.}
    \label{fig:motivation}
    \vspace{10pt}
\end{figure}

We choose to operate on a stereo setting since the monocular
case is substantially more complex as it requires to deal with
proper initialization of the features and has to handle scale drift.
Coping with these two aspects would add additional complexity that
requires more advanced computer vision skills. Yet a stereo visual SLAM
system is sufficiently usable to engage the students to continue
learning in this field. Our system processes rectified and
undistorted stereoscopic images, thus preventing the programmer from
the need of handling the lens distortion and having to handle the
stereo camera geometry. This comes at the cost of a slightly lower
accuracy, but our experiments show that despite this simplification
one can still achieve a high performance.

Similar to ORB-SLAM, our approach is feature based: it tracks a few
features in the scene and thanks to the known geometry of the stereo
cameras it determines the 3D position of the corresponding points.
The image points tracked along multiple subsequent frames are grouped
to form landmarks, that are salient points in the 3D spaces
characterized by an appearance. The landmarks observed along a small
portion of the trajectory are grouped in small point clouds (local
maps) and the local maps themselves are arranged in a pose graph. This
pose-graph~\cite{grisetti10titsmag} offers a deformable spatial
backbone for the local maps that can be altered whenever the robot
revisits a known location and identifies the current local map similar to an old one.
Albeit the different modules of our system could be easily
parallelized, we suggest a single threaded implementation, thus
avoiding the complexity induced by synchronizing multiple threads and
preserving the integrity of the memory.  

Yet with this straightforward non parallel pipeline and with other
further simplifications, such as the absence of any bundle adjustment
stage in the system, our approach achieves accuracy comparable to the
one of state of the art algorithms. Furthermore \ourmethod~has
substantially lower computational requirements. We performed
comparative experiments on standard benchmarking datasets acquired from heterogeneous platforms, namely
cars and quadrotors. 
Figure~\ref{fig:motivation} shows the outcome of our approach while processing
standard KITTI and EuRoC datasets acquired with different platforms
Finally, we contribute to the community by providing an overseeable yet complete open-source
SLAM system which can compete with the state of the art.

\section{Related Work}
\label{sec:related}



In the remainder of this section we discuss a selection of large-scale stereo visual SLAM systems
and highlight their similarities and differences with respect to our approach.
One of the first online large-scale stereo visual SLAM that appeared in literature is FrameSLAM.
Konolige and Agrawal~\cite{konolige2008frameslam} introduced a complete feature based SLAM system with bundle adjustment
running in real-time. For their approach they use CenSure features and integrate IMU information into the odometry computation.
FrameSLAM proposed similar notions of system components like the ones of our system.

Pire \emph{et al.}~\cite{pire2015stereo} presented a compact, appearance based stereo visual method S-PTAM.
S-PTAM runs on 2 threads in real-time using the same BRIEF features as we do for tracking. 
Where g2o~\cite{kummerle2011g} is used for full bundle adjustment.
In contrast to the other presented algorithms S-PTAM is not performing explicit relocalization and relies on
full bundle adjustment for preserving the map consistency thus resulting in growing complexity as the size of the mapping environment increases.

Mur-Artal and Tardos~\cite{mur2016orb} recently introduced an excellent, open source stereo visual SLAM system they named ORB-SLAM2.
The system originated from the prominent, monocular ORB-SLAM published by the same authors.
ORB-SLAM2 achieves extraordinary performance on standard datasets thanks to a highly reliable tracking front end
and frequent relocalization using ORB features.
Mur-Artal defines compact objects some of which directly correspond to objects in \ourmethod~(e.g. landmarks).
The system manages to close loops in real-time by utilizing a bag of words approach first proposed by Galvez-Lopez and Tardos~\cite{galvez2012bags}.
ORB-SLAM2 employs g2o~\cite{kummerle2011g} for local bundle adjustment.
The ORB-SLAM2 pipeline is designed to run on 3 parallel threads, increasing the complexity of the system layout.

Stereo LSD-SLAM proposed by Engel \emph{et al.}~\cite{engel2015large} is a direct, featureless SLAM approach operating in large-scale
at high processing speeds faster than real-time on a single thread.
Engel exploits static and temporal stereo image changes at pixel level while also considering lightning changes.  

In contrast to these approaches, that aim at advancing the state of the art at the cost of increasing complexity,
\ourmethod~is designed to be easy to understand and implement.
Yet our system achieves comparable accuracy and has equal or lower computational requirements.

\section{Our Approach}\label{sec:approach}



The goal of \ourmethod~is to process sequences of stereo image pairs to generate a 3D \emph{map}.
This map should represent the environment perceived by the robot and support crucial functionalities required in a SLAM system.
The basic geometric entity that constitutes a map is a \emph{landmark}.
A landmark is a salient 3D point in the world characterized by its appearance in all images that display the landmark.
The appearance is captured by a descriptor so that landmarks that appear similar will also have a similar descriptor.

Landmarks acquired in a nearby region form a \emph{local map}, which can be seen as a point cloud where each point (landmark) has multiple descriptors.
The local maps are arranged spatially in a \emph{pose graph}.
Each node of a pose graph thus encodes a 3D isometry (rotation and translation),
representing the pose of the corresponding local map in the world.
Edges between local maps represent spatial constraints correlating local maps close in space.
These constraints are generated either by \emph{tracking} the camera motion between temporally subsequent local maps
or by aligning local maps acquired at distant times as a consequence of \emph{relocalization} events.

The core functionalities of a SLAM map are:
\begin{itemize}
\item Relocalization
\item Adjustment
\end{itemize}
Relocalization is achieved by comparing descriptors of local maps.
Arranging the local maps in a pose graph allows us to utilize existing factor graph optimization engines for adjustment.
Furthermore they enable us to limit the size of the adjustment problem,
compared to a full bundle adjustment approach, substantially reducing computational cost.
This comes at the price of a loss in accuracy, however our experiments show that
the approach presented in this paper still reaches state of the art performance.

\begin{figure}[ht]
	\centering
	\includegraphics[width=0.99\columnwidth]{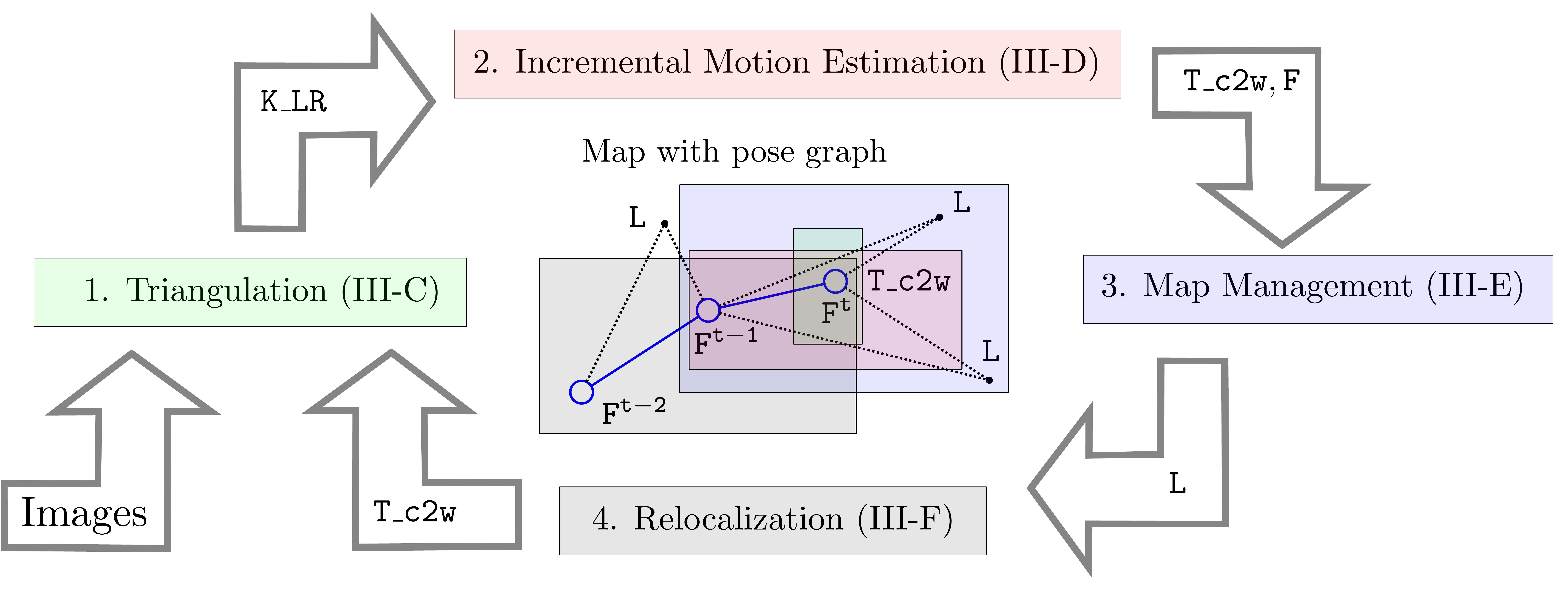}
	\caption{\ourmethod~system overview with the 4 core modules. The only external input to the system are stereo images.
	Differently shaded areas of the central map show the components accessed by the respective modules.}
	\label{fig:system_overview}
	\vspace{0pt}
\end{figure}

The process of generating these local maps can be split in 4 self-contained modules
illustrated in \figref{fig:system_overview}.
Their main tasks executed in sequence are:
\begin{itemize}
\item Triangulation (\ref{sec:triangulation}) - 
The Triangulation module takes a stereo pair of images as input
and produces 3D points plus corresponding feature descriptors for the left and right image respectively.
The collection of this output is stored in a structure which we name \emph{frame}.
\item Incremental Motion Estimation (\ref{sec:motion_estimation}) - 
In the Incremental Motion Estimation module we process a pair of subsequent frames
to subsequently obtain the pose of the current frame.
\item Map Management (\ref{sec:map_management}) - 
The Map Management module consumes frames with known camera motion and generates local maps
extending the pose graph and refining landmark positions.
\item Relocalization (\ref{sec:relocalization}) - 
Relocalization is done by seeking if the current local map appears similar to
some other local map generated in the past.
Upon a successful search we derive the spatial relations between them and update our entire world map.
\end{itemize}

In the following section we present an integrated overview over all major data structures of our system.
We make use of C++ like pseudocode notation for introducing our data structures and code snippets to encourage direct re-usability.

\subsection{Data Structures}\label{sec:datastructures}

The sole inputs to our system are rectified, undistorted stereoscopic intensity images.
Such \emph{Images} are represented by a pair of 2D arrays containing intensity values
and will be referred to as:
\begin{eqnarray}
\label{eq:datastructure_images}
  \mathtt{Image~I\_L},~\mathtt{I\_R}~\text{with}~\mathtt{I[r][c]} \in [0, 1]. \notag
\end{eqnarray}
Where the subscripts $\mathtt{\_L}$ and $\mathtt{\_R}$ refer to the left respectively right camera of the stereo configuration.
The integers $\mathtt{r/c}$ are equivalent to the pixel row/column indices of an image,
also referred to as image coordinates.

We further introduce a more complex structure holding a feature's keypoint and descriptor information:
\snugshade\begin{lstlisting}
KeypointWD {
  int r;          //image row of keypoint location
  int c;          //image column of keypoint location
  float response; //keypoint response
  Descriptor d;   //corresponding descriptor
};
\end{lstlisting}\endsnugshade
A \emph{KeypointWD} (WD for WithDescriptor) $\mathtt{K}$ is created upon Feature Detection (\ref{sec:triangulation}-1).
The corresponding descriptor value $\mathtt{K.d}$ is set after Descriptor Extraction (\ref{sec:triangulation}-3).
The value $\mathtt{K.response}$ represents the 'goodness' of a feature, a higher value indicating a more reliable redetection.

For these keypoints $\mathtt{K}$ we can retrieve epipolar correspondences to recover geometric information.
The collection of data obtained during this process is stored in a so-called \emph{framepoint} $\mathtt{P}$:
\snugshade\begin{lstlisting}
Framepoint {
  KeypointWD k_L, k_R;     //stereo keypoint+descriptor
  Vector3 p_c;             //3D coordinates in camera
  Vector3 p_w;             //3D coordinates in world
  Framepoint *prev, *next; //track information
  Landmark *landmark;      //linked landmark
  bool inlier;             //pose opt. inlier status
};
\end{lstlisting}\endsnugshade
By matching framepoints from the current image $\{\mathtt{P_t}\}$ against the ones from the previous image $\{\mathtt{P_{t-1}}\}$ we can determine
which framepoints potentially originate from the same location in the real world.
Once we found two such framepoints we link them by setting $\mathtt{P_t.prev}=\mathtt{P_{t-1}}$.
These chains of framepoints are also called \emph{tracks} and are derived inside the Incremental Motion Estimation module~(\ref{sec:motion_estimation}).

The Triangulation module usually manages to detect hundreds to thousands of good stereo point candidates in an Image pair,
resulting in an approachingly high number of framepoints.
The produced framepoints are accumulated in a frame $\mathtt{F}$:
\snugshade\begin{lstlisting}
Frame {
  Image I_L, I_R;         //raw image data
  Transform T_c2w, T_w2c; //pose in world
  Framepoint points[];    //present framepoints (owned)
  Camera *cam_L, *cam_R;  //cameras (stereo configuration)
};
\end{lstlisting}\endsnugshade
In addition to the framepoints $\mathtt{F.points}$, a frame also holds the orientation
and position estimate of the camera to the world $\mathtt{T\_c2w}$ for the given images $\mathtt{I\_L,R}$ at the time of its creation.
The pose $\mathtt{T\_c2w}$ is also referred to as the frame pose and is always expressed respective to the world coordinate frame.
Additionally a frame references a left and a right camera object, both 
capable of projecting framepoints~\eqref{eq:framepoint_projection} into an Image $\mathtt{I}$.

We further introduce the landmark structure $\mathtt{L}$:
\snugshade\begin{lstlisting}
Landmark {
  Vector3 p_w;        //landmark position in world
  Framepoint *origin; //first sight of this landmark
  Matrix3 omega;      //information matrix
  Vector3 nu;         //information vector
};
\end{lstlisting}\endsnugshade
Landmarks combine the information of a track of framepoints into a single entity.
A landmark has one timeless position $\mathtt{L.p\_w}$, describing its currently estimated coordinates in the world frame.
Using the $\mathtt{L.origin}$ field one has access to the first linked framepoint.
The framepoint fields $\mathtt{P.next}$ and $\mathtt{P.prev}$ allow for easy navigation through the complete track history.
In essence, landmarks can seen as framepoints with temporal information and are the closest property to a real world point.
We favor landmarks over framepoints in two modules of our pipeline (\ref{sec:motion_estimation}, \ref{sec:relocalization}).
Since landmarks exist over a sequence of images they are less affected by motion drift than framepoints
and thus provide a more stable anchor for position estimation.
The properties $\mathtt{omega}$ and $\mathtt{nu}$ are storing measurement information
which is used in the information filter of the Landmark optimization phase.

For \ourmethod, we chose to bundle a set of subsequent frames together to perform reliable and fast relocalization (\ref{sec:relocalization}).
This bundling results in a local map object $\mathtt{M}$:
\snugshade\begin{lstlisting}
LocalMap {
  Transform T_c2w, T_w2c; //pose of the local map in world
  Frame frames[];         //enclosed frames (owned)
  Landmark landmarks[];   //enclosed landmarks (shared)
};
\end{lstlisting}\endsnugshade
A local map object $\mathtt{M}$ carries a transform information $\mathtt{M.T\_c2w}$, defined by the last frame in the collection of $\mathtt{M.frames}$.
Additionally $\mathtt{M}$ holds a collection of direct references to the contained landmarks $\mathtt{L}$.
Subsequently we define our global map object to be:
\snugshade\begin{lstlisting}
WorldMap {
  Transform T_c2w, T_w2c; //world map origin
  LocalMap maps[];        //enclosed local maps (owned)
  Landmark landmarks[];   //enclosed landmarks (owned)
  OptimizableGraph graph; //pose graph (g2o)
};
\end{lstlisting}\endsnugshade
The world map defines the world coordinate frame for landmarks and local maps.
The contained pose graph is used by the relocalization module
to obtain optimized local map poses after adding a loop closure.

\begin{figure}[ht]
	\centering
	\includegraphics[width=0.99\columnwidth]{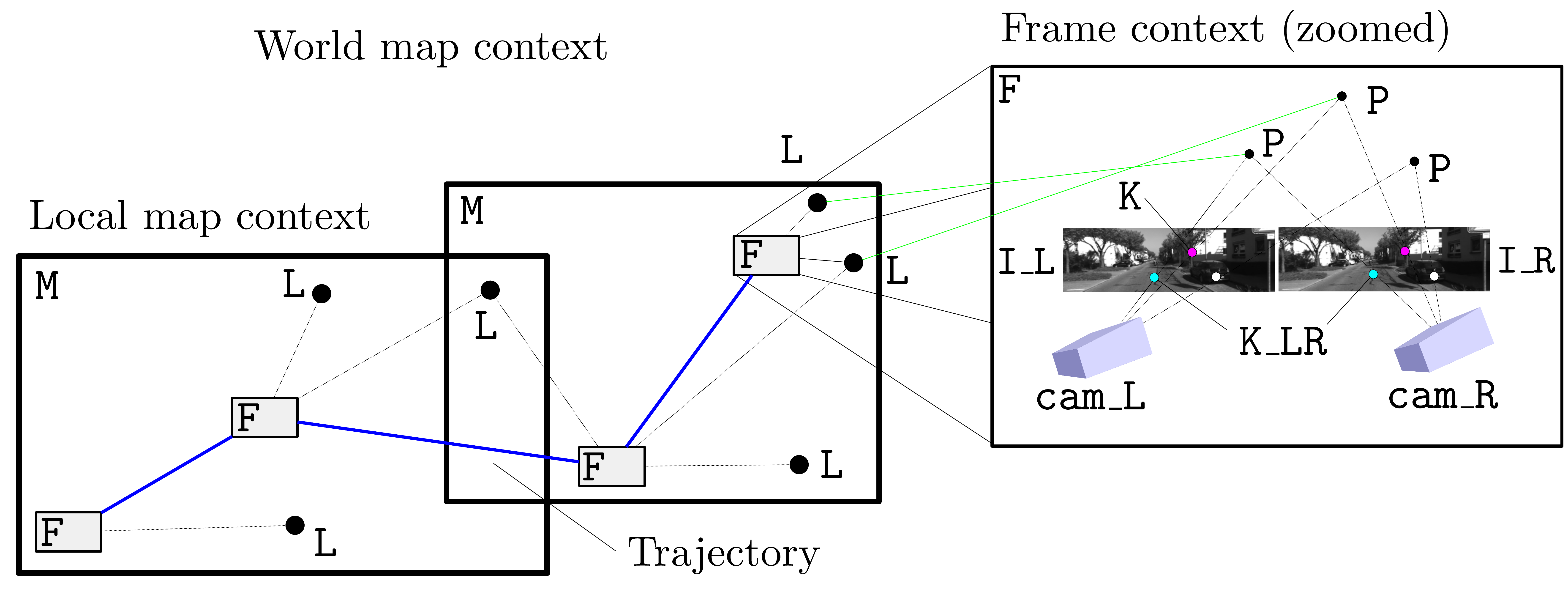}
	\caption{Schematic overview of our data structures.
	The trajectory is highlighted in blue, frames $\mathtt{F}$ and local maps $\mathtt{M}$ as rectangles and landmarks $\mathtt{L}$ as black dots.
	A stereo keypoint pair $\mathtt{K\_LR}$ (colored dots in images $\mathtt{I\_L}$, $\mathtt{I\_R}$) is visualized in the zoomed frame.
	The green lines connect framepoints $\mathtt{P}$ with their set landmark ($\mathtt{P.landmark}$).}
	\label{fig:datastructures}
	\vspace{0pt}
\end{figure}

\subsection{Triangulation}\label{sec:triangulation}

The goal of the Triangulation module is to process raw image data into a frame containing a cloud of framepoints.
We assume the standard pinhole camera model~\cite{hartley2003multiple} for all projection related operations.
In a first stage we perform Feature Detection, returning keypoints $\mathtt{K\_L}$, $\mathtt{K\_R}$.
To this extent we make use of the standard FAST keypoint detector~\cite{rosten2006machine}.
In order to maintain a steady number of returned keypoints the FAST threshold is adjusted automatically at runtime (e.g. to obtain less points the value is raised and vice versa).

In a next phase we ensure that the keypoints are evenly distributed among the left image.
The Regularization unit achieves this by first dividing all keypoints $\mathtt{K\_L}$ into bins arranged as a fine grid on the image.
Subsequently only the keypoint with the highest $\mathtt{K.response}$ value in each bin is kept while discarding all the other keypoints.
This operation heavily reduces the number of keypoints in the left image.
On the right image we do not perform regularization in order to preserve the maximum number of potential candidates for the stereo point triangulation.

During the Descriptor Extraction the feature descriptors $\mathtt{K.d}$ are computed and set to $\mathtt{K\_L}$, $\mathtt{K\_R}$.
For this purpose we use standard BRIEF descriptors~\cite{calonder2010brief}, 
being a decent choice for real-time applications thanks to their robust and lightweight architecture.

Having $\mathtt{K\_L}$, $\mathtt{K\_R}$ with set descriptors allows us to perform an epipolar search for stereo keypoint pairs $\mathtt{K\_LR}$.
This process is significantly simplified since we are working with undistorted and rectified images.
In \lstref{lst:line_search} we depict the implementation of the exact search procedure.
\snugshade\begin{lstlisting}[caption={Stereo keypoint search},label=lst:line_search]
//Require: input arrays from left and right image
KeypointWD K_L[] = .., K_R[] = ..;
//Ensure: output array of stereo keypoint pairs
pair<KeypointWD, KeypointWD> K_LR[];
//sort all input vectors in the order of the expression
sort(K_L, ((K_L[i].r < K_L[j].r) || 
           (K_L[i].r == K_L[j].r && K_L[i].c < K_L[j].c)));
sort(K_R, ((K_R[i].r < K_R[j].r) || 
           (K_L[i].r == K_L[j].r && K_R[i].c < K_R[j].c)));
//configuration
const float maximum_matching_distance = ..;
int idx_R = 0;
//loop over all left keypoints
for (int idx_L = 0; idx_L < K_L.size(); idx_L++) {
  //stop condition
  if (idx_R == K_R.size()) {break;}
  //the right keypoints are on an lower row - skip left
  while (K_L[idx_L].r < K_R[idx_R].r) {
    idx_L++; if (idx_L == K_L.size()) {break;}
  }
  //the right keypoints are on an upper row - skip right
  while (K_L[idx_L].r > K_R[idx_R].r) {
    idx_R++; if (idx_R == K_R.size()) {break;}
  }
  //search bookkeeping
  int idx_RS      = idx_R;
  float dist_best = maximum_matching_distance;
  int idx_best_R  = 0;
  //scan epipolar line for current keypoint at idx_L
  while (K_L[idx_L].r == K_R[idx_RS].r) {
    //zero disparity stop condition
    if (K_R[idx_RS].c >= K_L[idx_L].c) {break;}
    //compute descriptor distance
    const float dist = hnorm(K_L[idx_L].d, K_R[idx_RS].d)
    if(dist < dist_best) {
      dist_best  = dist;
      idx_best_R = idx_RS;
    }
    idx_RS++;
  }
  //check if something was found
  if (dist_best < maximum_matching_distance) {
    K_LR += pair(K_L[idx_L], K_R[idx_best_R]);
    idx_R = idx_best_R+1;
  }
}
\end{lstlisting}\endsnugshade
For an image point in the left frame ($\mathtt{r\_L,c\_L}$) the corresponding image point in the right frame ($\mathtt{r\_R,c\_R}$)
must lie on a pixel column index ($\mathtt{c\_R}$) equal or less than the one from the left ($\mathtt{c\_L}$).
This allows us to devise an efficient search strategy by ordering the keypoint vectors
according to the lexicographical order of rows and columns (see \lstref{lst:line_search}).
Once the vectors are sorted the search can be done in linear time.
The $\mathtt{hnorm}$ function returns the \emph{Hamming distance} between two descriptors.
Upon completion, the Stereo keypoint search returns a set of stereo keypoint pairs $\mathtt{K\_LR}$.

Ultimately the keypoint pairs $\mathtt{K\_LR}$ are triangulated in order to obtain 3D points $\mathtt{p\_c}$.
The computation of the depth $\mathtt{p\_c.z}$ based on a epipolar keypoint pair $(\mathtt{k\_L},\mathtt{k\_R})\in\mathtt{K\_LR}$ goes as follows:
\begin{eqnarray}
\label{eq:triangulation_depth}
  \mathtt{p\_c.z} = \mathtt{B}/(\mathtt{k\_R.c}-\mathtt{k\_L.c})
\end{eqnarray}
Where $\mathtt{B}$ is the stereo camera baseline (pixels$\times$meters) and $\mathtt{k\_L.c}$
the left horizontal image coordinate of a keypoint pair $(\mathtt{k\_L},\mathtt{k\_R})$.
Having the depth one can compute the $\mathtt{p\_c.x}$ and $\mathtt{p\_c.y}$ camera coordinates according to:
\begin{eqnarray}
\label{eq:triangulation_xy}
  \mathtt{p\_c.x} = \mathtt{p\_c.z}/\mathtt{F\_x}*(\mathtt{k\_L.c}-\mathtt{C\_x}) \\
  \mathtt{p\_c.y} = \mathtt{p\_c.z}/\mathtt{F\_y}*(\mathtt{k\_L.r}-\mathtt{C\_y})
\end{eqnarray}
Where $\mathtt{F\_x,y}$ are the focal lengths and $\mathtt{C\_x}$, $\mathtt{C\_y}$ the principal points (pixels) of the left cameras projection matrix.

\begin{figure}[ht]
	\centering
	\includegraphics[width=0.99\columnwidth]{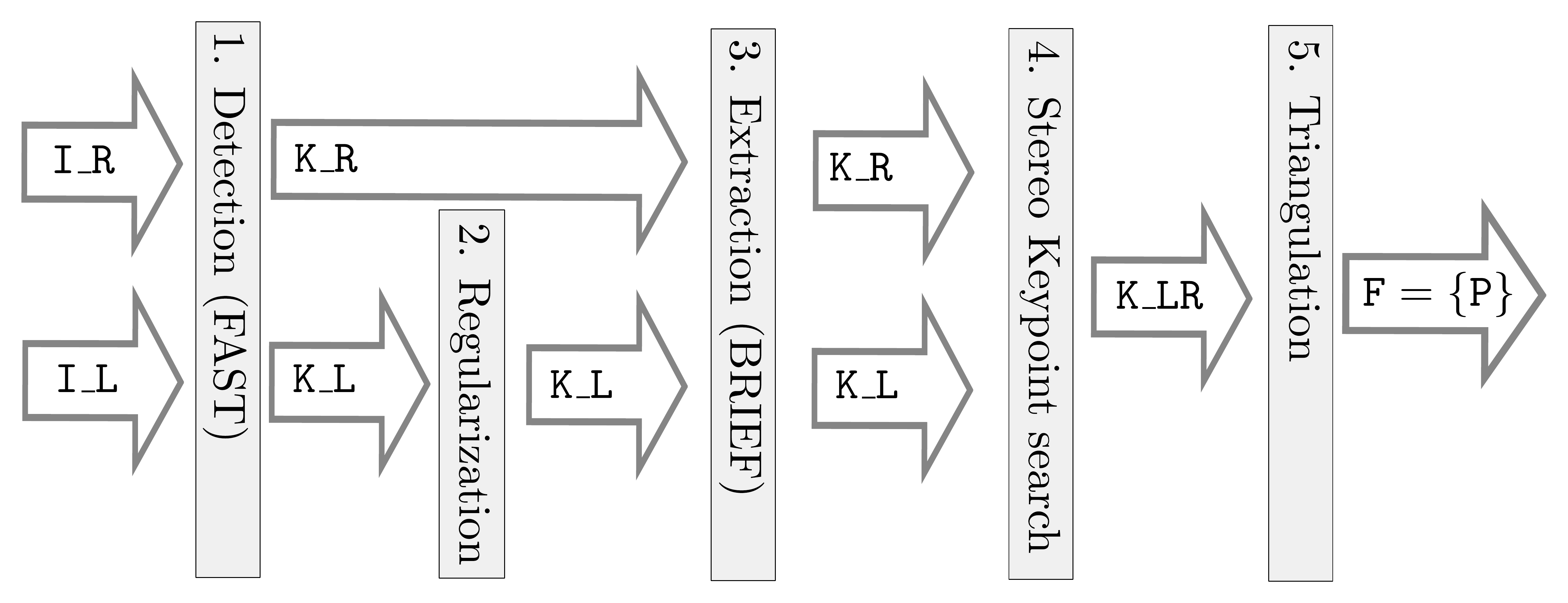}
	\caption{Triangulation (\ref{sec:triangulation}) module.
	From left to to right, raw image data is transformed into an augmented 3D point cloud $\{\mathtt{P}\}$ contained in a frame object $\mathtt{F}$.
	Note that the keypoints from the right image are not filtered by the Regularization process.}
	\label{fig:triangulation}
	\vspace{0pt}
\end{figure}

Once all keypoint pairs have been processed, the module packages each pair $(\mathtt{k\_L},\mathtt{k\_R})$ with its point $\mathtt{p\_c}$ into a framepoint $\mathtt{P}$.
\figref{fig:triangulation} shows the entire Triangulation module with inputs, outputs and its processing units.

\subsection{Incremental Motion Estimation}\label{sec:motion_estimation}


The goal of the Motion Estimation module is to determine the relative motion the left camera underwent, to get
from a previous frame $\mathtt{F_{t-1}}$ to the current frame $\mathtt{F_t}$
in order to locate the current frame.
That motion $\mathtt{T\_m}$ can be inferred as the same motion framepoints from the previous frame took, to appear in the current.
Meaning that if we manage to connect framepoints in $\mathtt{F_t}$
with framepoints in $\mathtt{F_{t-1}}$ which stem from the same world point,
we can compute the relative transform between them and subsequently find the pose $\mathtt{T\_c2w}$ of our current frame.

The task of finding these framepoint correspondences (tracks) is handled by the Tracking unit.
We seek to find tracks based on projected image coordinates on the left camera only (monocular tracking).
For this endeavor we first have to get the previous framepoints into the current image.
A framepoint $\mathtt{P_{t-1}}$ from the previous frame is projected into the current left image $\mathtt{I\_L}$ as:
\begin{eqnarray}
\label{eq:framepoint_projection}
 \mathtt{k\_P}=\pi(\mathtt{P\_L}*\mathtt{T\_w2c}*\mathtt{P_{t-1}.p\_w})
\end{eqnarray}
Where $\pi(\mathtt{p\_w})$ is the camera projection function~\cite{hartley2003multiple} and $\mathtt{P\_L}$ the projection matrix of the left camera.
The object $\mathtt{k\_P}$ can be interpret as a new keypoint, carrying the descriptor of the framepoint $\mathtt{P.k\_L.d}$.
To obtain the initial pose estimate $\mathtt{T\_w2c}$ we use the \emph{constant velocity motion model}:
\begin{eqnarray}
\label{eq:motion_model}
  \mathtt{T\_m}&=&\mathtt{F_{t-1}.T\_w2c}*\mathtt{F_{t-2}.{(T\_w2c)}^{-1}} \\ \notag
  \mathtt{T\_w2c}&=&\mathtt{T\_m}*\mathtt{F_{t-1}.T\_w2c}
\end{eqnarray}
If the frame $\mathtt{F_{t-2}}$ is not available we assume no motion and the prior becomes: $\mathtt{T\_w2c}=\mathtt{F_{t-1}.T\_w2c}$.
This simple motion prediction model is sufficiently robust for systems with smooth motion and decent frame rates.

For each predicted keypoint $\mathtt{k\_P}$ from the previous frame we search for the keypoint $\mathtt{k\_L}$ in
the current image that is closest to $\mathtt{k\_P}$.  We restrict the search to a rectangular region around 
$\mathtt{k\_P}$ and we consider only those keypoint whose descriptors $\mathtt{k\_L.d}$ appear similar enough 
to $\mathtt{k\_P.d}$.
The search is completed once each projection $\mathtt{k\_P}$ is matched to a keypoint $\mathtt{k\_L}$
or terminated if there are no more keypoints $\{\mathtt{k\_L}\}$ available to match to.

\begin{figure}[ht]
	\centering
	\includegraphics[width=0.99\columnwidth]{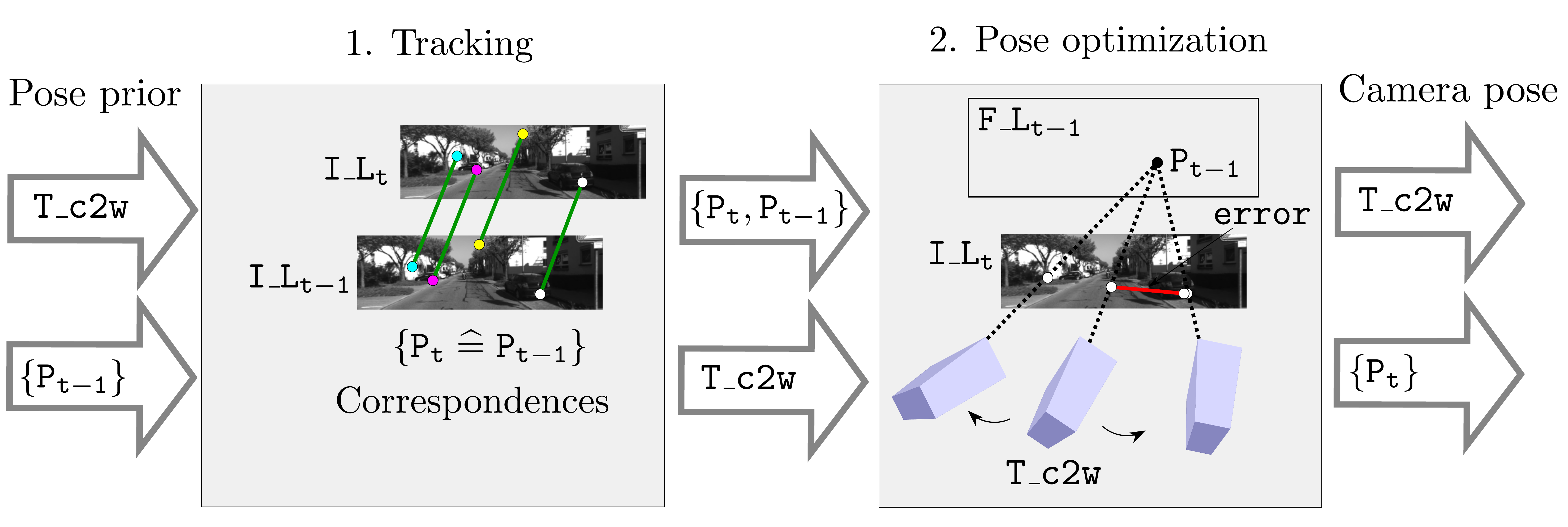}
	\caption{Incremental Motion Estimation (\ref{sec:motion_estimation}) module. Overview of the incremental
	motion estimation for one timestep from previous Image $\mathtt{I\_L_{t-1}}$ to current Image $\mathtt{I\_L_t}$.
	The output framepoints $\{\mathtt{P_t}\}$ with corresponding previous framepoints $\{\mathtt{P_{t-1}}\}$
	are connected by the $\mathtt{P.prev}$ field.}
	\label{fig:motion_estimation}
	\vspace{0pt}
\end{figure}

Given this set of matches we are able to determine the current camera pose.
The Pose optimization utilizes the pose prior $\mathtt{T\_w2c}$, used to determine the keypoint matches,
as initial guess and computes a refined estimate by minimizing the resulting projection errors in the image plane.
Our complete Pose optimization algorithm is presented in \lstref{lst:posit}.
\snugshade\begin{lstlisting}[caption={Pose optimization},label=lst:posit]
//Require: current frame (carrying framepoints)
Frame* frame = ..;
//Require|Ensure: transform estimate camera to/from world
Transform T_c2w = .., T_w2c = ..;
//ds configuration
const bool ignore_outliers       = ..;
const float kernel_maximum_error = ..;
const float close_depth          = ..;
const float maximum_depth        = ..;
const int number_of_iterations   = ..;
for (int i = 0; i < number_of_iterations; i++) {
  //initialize least squares components
  Matrix6 H = 0, Vector6 b = 0, Matrix4 omega = I;
  //loop over all framepoints
  for (Framepoint* fp: frame->points()) {
    if (!fp->prev()) {continue;}
    CameraCoordinates p_c = T_w2c*fp->prev()->p_w;
    //preferably use landmark position estimate
    if (fp->landmark()) {
      p_c   = T_w2c*fp->landmark()->p_w;
      //increase weight for landmarks
      omega = ..;
    }
    //project point into left and right image
    const Keypoint k_L = frame->cam_L->project(p_c);
    const Keypoint k_R = frame->cam_R->project(p_c);
    //ds compute current reprojection error
    const Vector4 error(k_L.u-fp->k_L.u, k_L.v-fp->k_L.v, 
                        k_R.u-fp->k_R.u, k_R.v-fp->k_R.v);
    //compute squared error
    const float error_squared = error.transpose()*error;
    //robust kernel
    if (error_squared > kernel_maximum_error) {
      if (ignore_outliers) {continue;}
        omega *= kernel_maximum_error/error_squared;
    } else {
      fp->inlier = true;
    }    
    //compute stereouv jacobian see Eq. (11)
    Matrix4_6 J = getJacobian(T_w2c, p_c, frame);
    //adjust omega: if the point is considered close
    if(p_c.z() < close_depth) {
      omega *= (close_depth-p_c.z())/close_depth;
    } else {
      omega *= (maximum_depth-p_c.z())/maximum_depth;
      //disable contribution to translational error
      J.block<3,3>(0,0) = 0;
    }            
    //update H and b
    H += J.transpose()*omega*J;
    b += J.transpose()*omega*error;                   
  }
  //compute (Identity-damped) solution
  const Vector6 dx = solve((H+I)\(-b));
  T_w2c = v2t(dx)*T_w2c;
  T_c2w = T_w2c.inverse();
}
\end{lstlisting}\endsnugshade
Note that the Pose optimization algorithm checks if framepoints are connected to landmarks and if so, 
uses the landmark position $\mathtt{L.p\_w}$ instead of the framepoint position $\mathtt{P.p\_w}$ to compute the reprojections $\mathtt{k\_L}, \mathtt{k\_L}$.
After each iteration the current pose estimate $\mathtt{T\_c2w}$ is updated.
The function $\mathtt{v2t}$ transforms the pose matrix from the smooth manifold representation~\cite{smith1986representation} back to an isometry.
For completeness we provide the complete stereo image point Jacobian:
\begin{eqnarray}
\label{eq:jacobian_stereoposit}
\mathtt{J\_L} &=&\left[\begin{array}{ccc} \mathtt{1/c\_L} & \mathtt{0} & -\mathtt{a\_{L}/c\_{L}^2} \\ 
\mathtt{0} & \mathtt{1/c\_{L}} & -\mathtt{b\_{L}/c\_{L}^2} \end{array}\right] \\
\mathtt{J\_R} &=&\left[\begin{array}{ccc} \mathtt{1/c\_R} & \mathtt{0} & -\mathtt{a\_{R}/c\_{R}^2} \\ 
\mathtt{0} & \mathtt{1/c\_{R}} & -\mathtt{b\_{R}/c\_{R}^2} \end{array}\right] \\
\mathtt{J\_T} &=& \left[\begin{array}{cc} \mathtt{I} & \mathtt{-2\mathtt * skew(p\_c)} \\
\mathtt{0}~\mathtt{0}~\mathtt{0} & \mathtt{0}~\mathtt{0}~\mathtt{0} \end{array}\right] \\
\mathtt{J} &=&\left[\begin{array}{c}
\mathtt{J\_L}*\mathtt{P\_L} \\
\mathtt{J\_R}*\mathtt{P\_R} \end{array}\right] *\mathtt{J\_T}
\end{eqnarray}
The coefficients $\mathtt{a\_L,R}$, $\mathtt{b\_L,R}$ and $\mathtt{c\_L,R}$ are the coordinates of $\mathtt{p\_c}$
in the left respective right image before the homogenous division.
$\mathtt{P\_L}$ and $\mathtt{P\_R}$ stand for the left and right camera projection matrix. In this particular case $\mathtt{I}$ stands for the $3\times3$ identity matrix.
\figref{fig:motion_estimation} shows an overview of the whole Incremental Motion Estimation module.
The final pose estimate $\mathtt{T\_c2w}$ and framepoints having field $\mathtt{P.inlier}$ set to $\mathtt{true}$
are passed to the next module.

\subsection{Map Management}\label{sec:map_management}

\begin{figure}[ht]
	\centering
	\includegraphics[width=0.99\columnwidth]{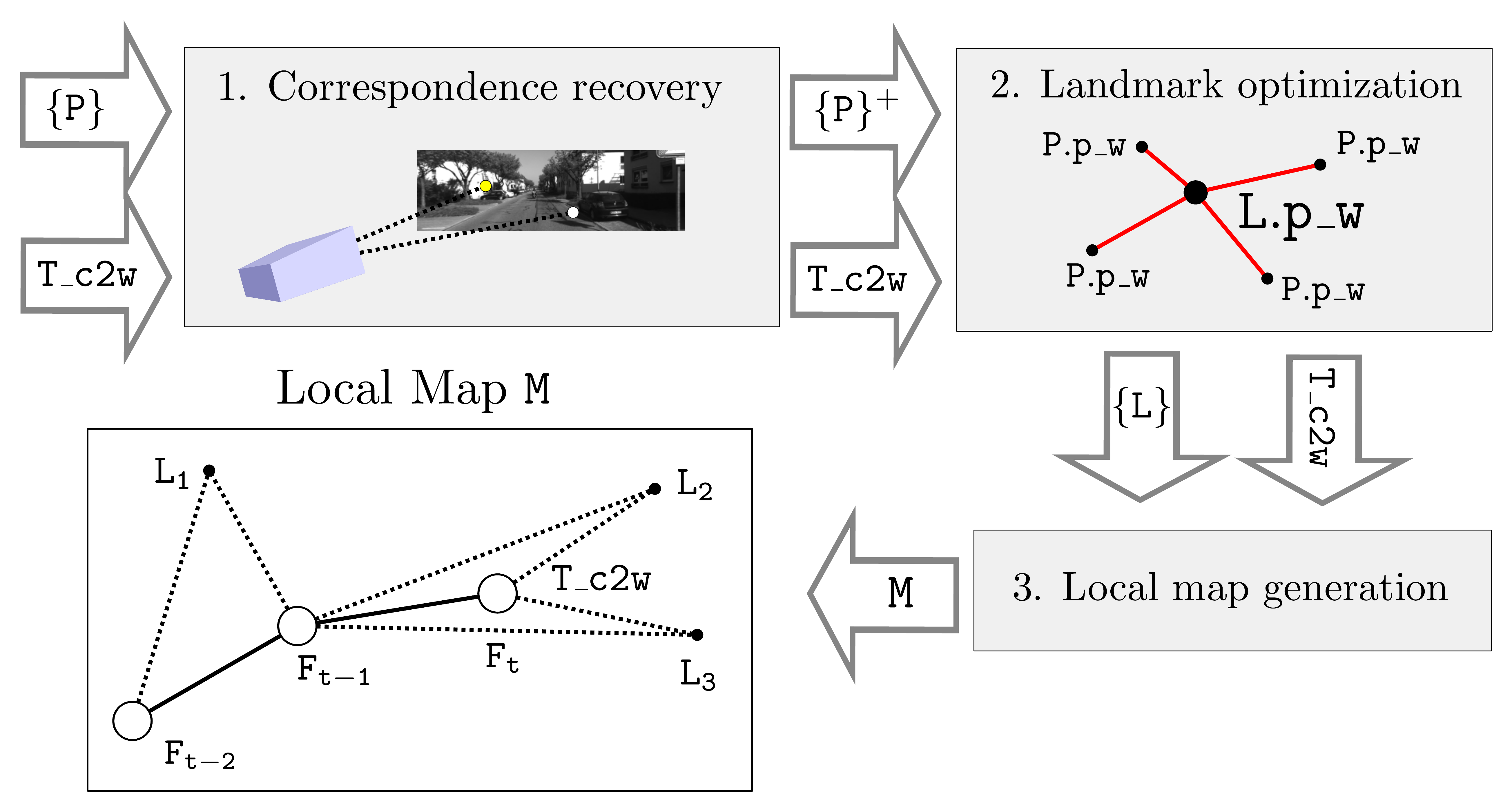}
	\caption{Map Management (\ref{sec:map_management}) module.
	Example sequence for the generation of a mocal map after 3 time steps (spanning over 3 frames).}
	\label{fig:map_management}
	\vspace{0pt}
\end{figure}

The Map Management module governs 3 main tasks in the following order:\vspace{5pt}
\begin{enumerate}
\item Correspondence recovery
\item Landmark optimization
\item Local map generation\vspace{5pt}
\end{enumerate}
Correspondence recovery is an approach to locate previous framepoints in the current image by using
the precise motion estimate resulting from the previous module (\ref{sec:motion_estimation}).
The set of candidates for Correspondence recovery is generated by bookkeeping of 
unmatched framepoints in the Tracking unit (\ref{sec:motion_estimation}-1) of the Motion Estimation module.
First we obtain the projection $\mathtt{k\_P}$ of the framepoint in the current image according to \eqref{eq:framepoint_projection}.
Then the descriptor at $\mathtt{k\_P}$ is compared against $\mathtt{P.d\_L}$.
If the matching distance is acceptable we confirm the correspondence and create a new framepoint linked to the previous one.
Each time a landmark is observed, its estimate is refined by using a running information filter.
A new local map is generated if one or both of the following properties exceed a certain threshold:
\begin{itemize}
\item Translation relative to previous local map
\item Rotation relative to previous local map
\end{itemize}
For the generation of a local map the current frame pose $\mathtt{F.T\_c2w}$ is used also for the local map pose $\mathtt{M.T\_c2w}$.
All other frames collected since the last Local map generation are scanned for landmarks and stored together in the new local map.
The complete Map Management module can be inspected in \figref{fig:map_management}.

\subsection{Relocalization}\label{sec:relocalization}

We want our system to recognize a previously traversed place and use this information to improve the position estimate of the current frame.
A straightforward solution to this problem would be to compare each new frame against all past frames
and attempt to retrieve a spatial constraint.

We chose to relocalize a current local map $\mathtt{M_i}$ only in past local maps $\mathtt{M_j}$, and not in single frames $\mathtt{F}$
to use the richness of the local map structure (e.g landmarks).
Relocalization candidates are retrieved by comparing entire descriptor clouds from landmarks of local maps.
With the goal of obtaining landmark to landmark correspondences.
Searching correspondences with a similarity search is an expensive operation.
Hence we do not directly look for correspondences but rather want to have a matching estimate
of how much two local maps $\mathtt{M_i}$, $\mathtt{M_j}$ overlap first.
And if this estimate is sufficiently high we perform the Similarity search.
For both of these task we utilize the Hamming binary search tree library HBST~\cite{schlegel2016visual}.

The HBST library performs similarity search directly on descriptor to descriptor correspondences for two sets of descriptors.
The search is performed efficiently by arranging the descriptor sets in a binary search tree, which is
constructed after performing a probabilistic analysis on the descriptor bits.
We use HBST's descriptor to descriptor correspondences
to retrieve landmark to landmark correspondences ($\mathtt{L_{M_i}}\mathrel{\widehat{=}}\mathtt{L_{M_j}}$).
On $\mathtt{L_{M_i}}\mathrel{\widehat{=}}\mathtt{L_{M_j}}$ we run classic ICP~\cite{besl1992method} to achieve an Alignment.
Giving us the relative transform $\mathtt{T\_i2j}$ between the local map poses $\mathtt{M_i}$, $\mathtt{M_j}$. 
The quality of the solution is defined by the final number of ICP inliers and the resulting average error.
We discard transforms with a low inlier count and high average error in the Validation phase.

\begin{figure}[ht]
	\centering
	\includegraphics[width=0.99\columnwidth]{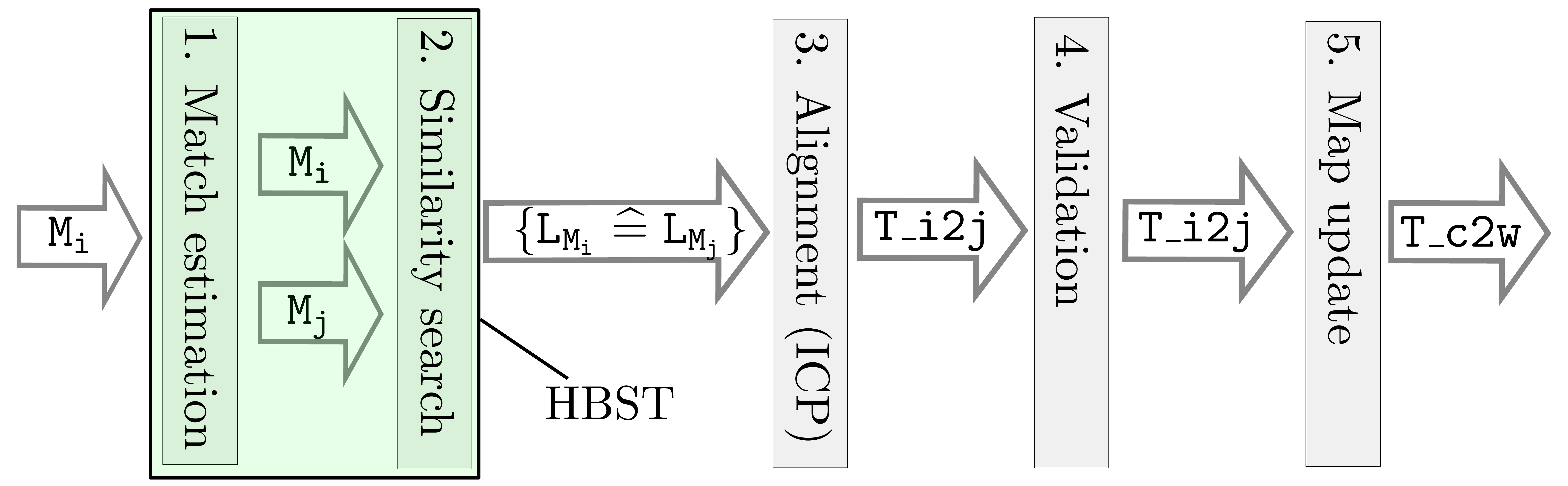}
	\caption{Relocalization (\ref{sec:relocalization}) module.
	The sequence can be prematurely escaped at any unit, resulting in the loss of a global map update.
	In that case the transform $\mathtt{T\_c2w}$ does not change its value.
	If the module achieves a successful relocalization the global map as well as $\mathtt{T\_c2w}$ gets refined.
	The highlighted section area the use of HBST functionality.}
	\label{fig:relocalization}
	\vspace{0pt}
\end{figure}

Validated relocalization constraints are introduced to the pose graph with a relaxed translational information value.
The pose graph is optimized using g2o~\cite{kummerle2011g}.
The optimized local map poses $\mathtt{M.T\_c2w}$ are broadcast to adjust the inner frame poses $\mathtt{F.T\_c2w}$.
Landmark positions $\mathtt{L.p\_w}$ are moved rigidly according to their last relative location in the respective local map.
\figref{fig:relocalization} shows the entire Relocalization module and its interaction with the HBST library.

\section{Experimental Evaluation}
\label{sec:exp}

The main focus of this work is to give the reader a clear insight into a complete, yet lightweight SLAM system while providing
directly applicable code snippets accompanied by a solid performance confirmation. 

All of our experiments are designed to show the capabilities of our system and to support our key proposals,
such as having a functional, straightforward SLAM system running in a single thread.
Which is capable of competing with state of the art algorithms in achieved precision while excelling at processing speed on ground vehicles
and also works on airborne platforms.

We perform our evaluations on publicly available standard benchmark datasets. 
The KITTI SLAM evaluation presented by Geiger \emph{et al.}~\cite{geiger2012we} features challenging ground vehicle sequences in urban environments.
We also tested \ourmethod~on airborne vehicles in
the recently published EuRoC MAV dataset collection of Burri \emph{et al.}~\cite{burri2016euroc}.
Our system runs with identical parameters throughout all of these experiments.


\subsection{Trajectory Precision}

The first experiment is designed to measure the precision of our approach.
We report a quantitative as well as qualitative insights into the full KITTI evaluation,
featuring 11 diverse mapping scenarios.

\begin{figure}[ht]
	\centering
  \begin{subfigure}[t]{0.99\columnwidth}
     \resizebox{0.99\columnwidth}{!}{
\begin{picture}(7200.00,5040.00)%
    \gdef\gplbacktext{}%
    \gdef\gplfronttext{}%
    \gplgaddtomacro\gplbacktext{%
      \csname LTb\endcsname%
      \put(814,704){\makebox(0,0)[r]{\strut{}$0$}}%
      \csname LTb\endcsname%
      \put(814,1439){\makebox(0,0)[r]{\strut{}$0.5$}}%
      \csname LTb\endcsname%
      \put(814,2174){\makebox(0,0)[r]{\strut{}$1$}}%
      \csname LTb\endcsname%
      \put(814,2909){\makebox(0,0)[r]{\strut{}$1.5$}}%
      \csname LTb\endcsname%
      \put(814,3644){\makebox(0,0)[r]{\strut{}$2$}}%
      \csname LTb\endcsname%
      \put(814,4379){\makebox(0,0)[r]{\strut{}$2.5$}}%
      \put(1434,484){\makebox(0,0){\strut{}00}}%
      \put(1922,484){\makebox(0,0){\strut{}01}}%
      \put(2410,484){\makebox(0,0){\strut{}02}}%
      \put(2898,484){\makebox(0,0){\strut{}03}}%
      \put(3386,484){\makebox(0,0){\strut{}04}}%
      \put(3875,484){\makebox(0,0){\strut{}05}}%
      \put(4363,484){\makebox(0,0){\strut{}06}}%
      \put(4851,484){\makebox(0,0){\strut{}07}}%
      \put(5339,484){\makebox(0,0){\strut{}08}}%
      \put(5827,484){\makebox(0,0){\strut{}09}}%
      \put(6315,484){\makebox(0,0){\strut{}10}}%
    }%
    \gplgaddtomacro\gplfronttext{%
      \csname LTb\endcsname%
      \put(176,2541){\rotatebox{-270}{\makebox(0,0){\strut{}Average relative translation error (\%)}}}%
      \put(3874,154){\makebox(0,0){\strut{}KITTI sequence number}}%
      \put(3874,4709){\makebox(0,0){\strut{}KITTI benchmark: Translation error}}%
      \csname LTb\endcsname%
      \put(5816,4206){\makebox(0,0)[r]{\strut{}LSD-SLAM}}%
      \csname LTb\endcsname%
      \put(5816,3986){\makebox(0,0)[r]{\strut{}ORB-SLAM2}}%
      \csname LTb\endcsname%
      \put(5816,3766){\makebox(0,0)[r]{\strut{}ProSLAM}}%
    }%
    \gplbacktext
    \put(0,0){\includegraphics{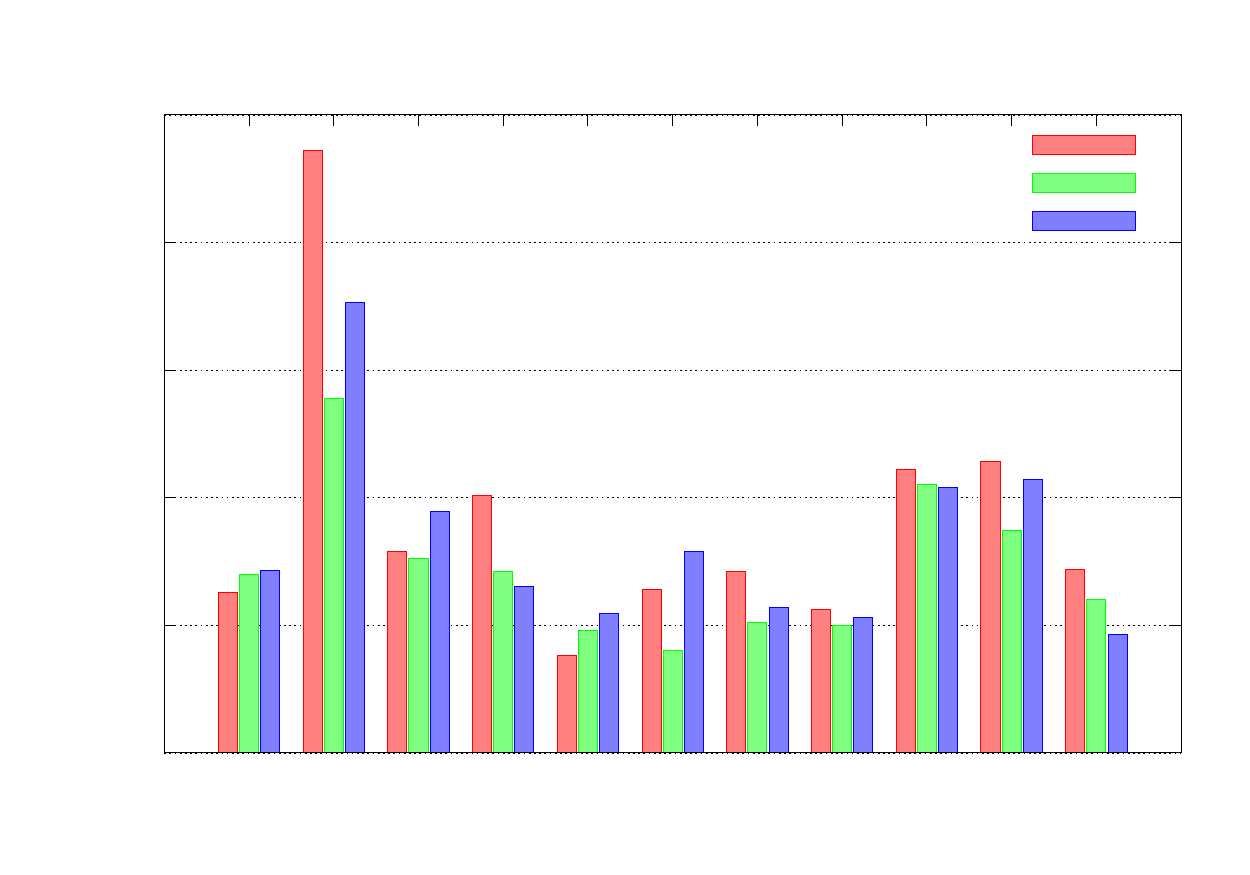}}%
    \gplfronttext
  \end{picture}}
     \label{fig:kitti_error_translation}
  \end{subfigure}
  \vspace{0pt}
	\begin{subfigure}[t]{0.99\columnwidth}
     \resizebox{0.99\columnwidth}{!}{
\begin{picture}(7200.00,5040.00)%
    \gdef\gplbacktext{}%
    \gdef\gplfronttext{}%
    \gplgaddtomacro\gplbacktext{%
      \csname LTb\endcsname%
      \put(814,704){\makebox(0,0)[r]{\strut{}$0$}}%
      \csname LTb\endcsname%
      \put(814,1439){\makebox(0,0)[r]{\strut{}$0.1$}}%
      \csname LTb\endcsname%
      \put(814,2174){\makebox(0,0)[r]{\strut{}$0.2$}}%
      \csname LTb\endcsname%
      \put(814,2909){\makebox(0,0)[r]{\strut{}$0.3$}}%
      \csname LTb\endcsname%
      \put(814,3644){\makebox(0,0)[r]{\strut{}$0.4$}}%
      \csname LTb\endcsname%
      \put(814,4379){\makebox(0,0)[r]{\strut{}$0.5$}}%
      \put(1434,484){\makebox(0,0){\strut{}00}}%
      \put(1922,484){\makebox(0,0){\strut{}01}}%
      \put(2410,484){\makebox(0,0){\strut{}02}}%
      \put(2898,484){\makebox(0,0){\strut{}03}}%
      \put(3386,484){\makebox(0,0){\strut{}04}}%
      \put(3875,484){\makebox(0,0){\strut{}05}}%
      \put(4363,484){\makebox(0,0){\strut{}06}}%
      \put(4851,484){\makebox(0,0){\strut{}07}}%
      \put(5339,484){\makebox(0,0){\strut{}08}}%
      \put(5827,484){\makebox(0,0){\strut{}09}}%
      \put(6315,484){\makebox(0,0){\strut{}10}}%
    }%
    \gplgaddtomacro\gplfronttext{%
      \csname LTb\endcsname%
      \put(176,2541){\rotatebox{-270}{\makebox(0,0){\strut{}Average rotation error (deg/100m)}}}%
      \put(3874,154){\makebox(0,0){\strut{}KITTI sequence number}}%
      \put(3874,4709){\makebox(0,0){\strut{}KITTI benchmark: Rotation error}}%
      \csname LTb\endcsname%
      \put(5816,4206){\makebox(0,0)[r]{\strut{}LSD-SLAM}}%
      \csname LTb\endcsname%
      \put(5816,3986){\makebox(0,0)[r]{\strut{}ORB-SLAM2}}%
      \csname LTb\endcsname%
      \put(5816,3766){\makebox(0,0)[r]{\strut{}ProSLAM}}%
    }%
    \gplbacktext
    \put(0,0){\includegraphics{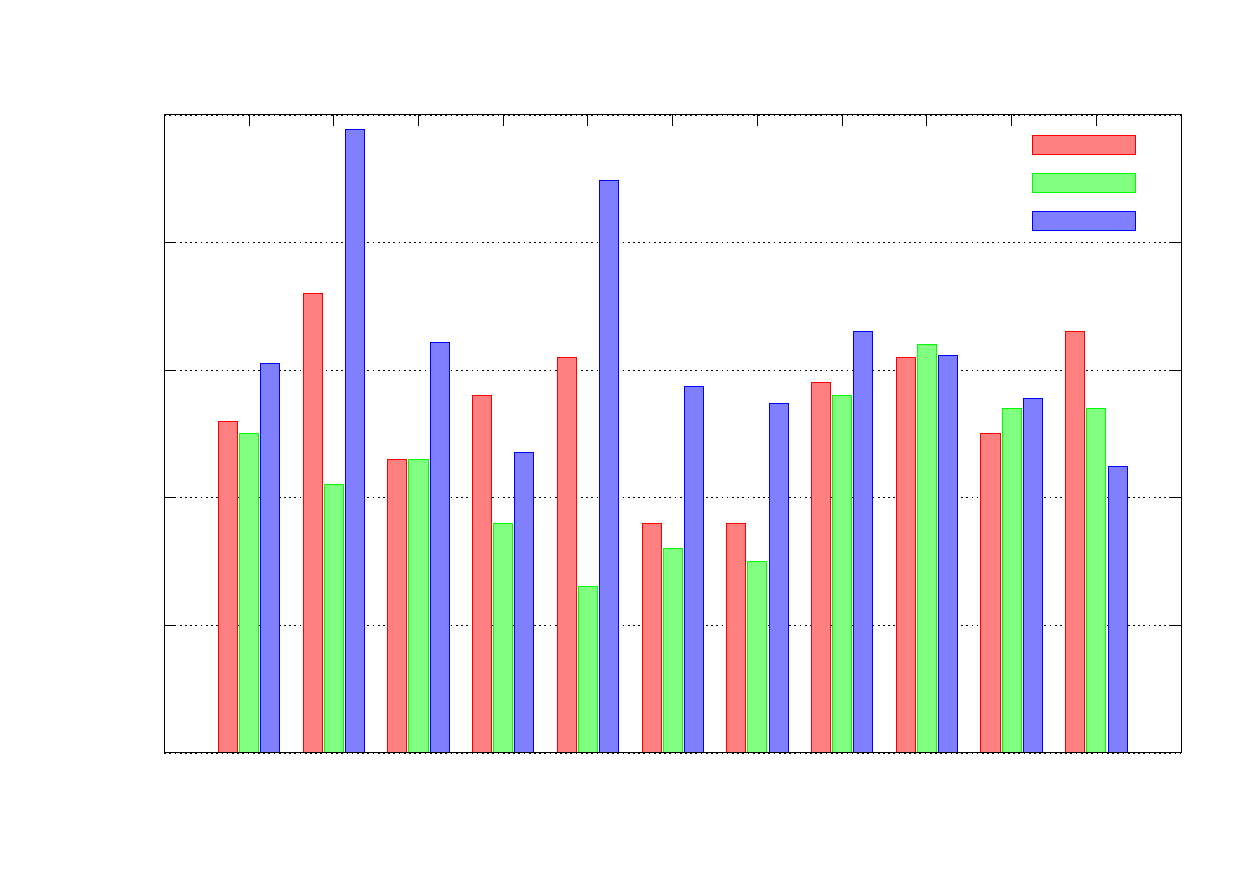}}%
    \gplfronttext
  \end{picture}}
     \label{fig:kitti_error_rotation}
  \end{subfigure}
	\caption{KITTI benchmark result comparison. Less is better.}
	\label{fig:kitti_benchmark}
	\vspace{0pt}
\end{figure}
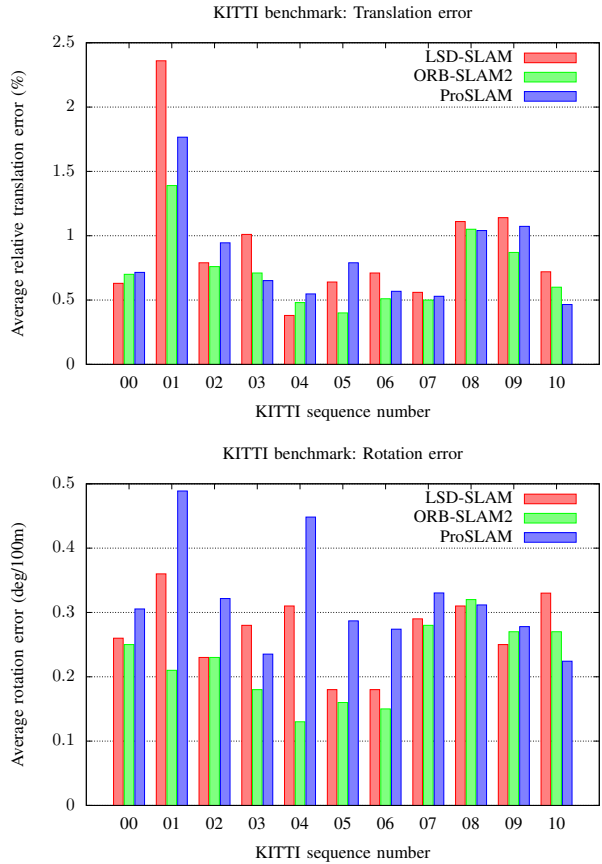

From the translation error we can tell that our approach stands in the comparable range with
the competing methods, outperforming LSD-SLAM in 7/11 and ORB-SLAM2 in 3/11 cases.
Note that all tested methods achieve a relative error of less than 1\% on average.
As for the rotation error, we see that our system performs weaker than any of the compared methods,
overcoming LSD-SLAM only in 2/11 and ORB-SLAM in 1/11 cases. 
We will address this issue in the Performance Analysis in~\secref{sec:performance_analysis}.

Additionally we tested our approach on data recorded from an airborne vehicle. 
In \tabref{tab:benchmark_euroc} we report the accuracy of our system in the EuRoC MAV Dataset using only the stereo stream.
We chose to display the absolute translation RMSE, being a clear indicator
whether a SLAM system manages to produce an accurate trajectory or not.

\begin{table}[ht]
  \caption{EuRoC benchmark result comparison (RMSE)}
  \centering
  {\footnotesize
  \begin{tabular}{|l|c|c|c|}
    \hline
    Dataset &LSD-SLAM & ORB-SLAM2 & \ourmethod \\
    \hline
    MH\_01\_easy & - & 0.035 & 0.087 \\
    MH\_02\_easy & - & 0.018 & 0.146 \\
    MH\_03\_medium & - & 0.028 & 0.272 \\
    MH\_04\_difficult & - & 0.119 & - \\
    MH\_05\_difficult & - & 0.060 & - \\
    V1\_01\_easy & 0.066 & 0.035 & 0.140 \\
    V1\_02\_medium & 0.074 & 0.020 & 0.211 \\
    V1\_03\_difficult & 0.089 & 0.048 & - \\
    V2\_01\_easy & - & 0.037 & - \\
    V2\_02\_medium & - & 0.035 & - \\
    V2\_03\_difficult & - & - & - \\
    \hline
  \end{tabular}
  }
  \label{tab:benchmark_euroc}
  \vspace{0pt}
\end{table}

The values for LSD-SLAM and ORB-SLAM2 correspond to the values reported in the publications~\cite{engel2015large,mur2016orb}.
For many challenging datasets our rudimentary motion model running on BRIEF features cannot adapt fast enough to the drones movement
in the dark or blurred areas and the track breaks.
The - sign indicates that the respective method was not able to finish the dataset.

\subsection{Performance Analysis}\label{sec:performance_analysis}

The experimental evaluation confirms that our system provides competitive results in trajectory precision.
\ourmethod~shows a weakness in rotation estimation compared to state of the art approaches.
This is a common odometry problem which is generally compensated by bundle adjustment.
We tolerate this issue since our system does not feature bundle adjustment and the rotation error is still very low.
On the other hand, we do not experience drift in pure translation due to the reliable depth values
we obtain from the stereo camera setup.
The overall system precision could have been further improved
by the integration of error propagation models in loop closing, however
to preserve clarity \ourmethod~does not use any error kind of error modeling.

\subsection{Processing Speed}

The next experiment was chosen to support the claim that our approach can be
executed fast enough to enable online processing on a single core CPU in real time.
We ran all 11 KITTI benchmark sequences 10 times for each method, capturing start and end time. 
The benchmark computer setup consisted of a portable computer equipped with:
\begin{itemize}
\item Ubuntu 16.04 and OpenCV 3.1.1
\item Intel i7-4700MQ CPU (4 cores, 3.4GHz, 6MB cache)
\end{itemize}
In the case of LSD-SLAM we report the values of the authors.
Considering that Engel used a more powerful Intel i7-4900MQ CPU (4 cores, 3.8GHz, 8MB cache)
for conducting the LSD-SLAM experiments~\cite{engel2015large}.

\begin{table}[ht]
  \caption{KITTI: Average image processing speed}
  \centering
  {\footnotesize
  \begin{tabular}{|l|c|c|c|c|}
    \hline
    Approach & Threads & Mean duration (s) & Std. Dev (s) \\
    \hline
    LSD-SLAM & 1 & 0.067 & 0.013 \\
    ORB-SLAM2 & 3 & 0.090 & 0.009 \\
    \ourmethod & 1 & \textbf{0.059} & 0.010 \\
    \hline
  \end{tabular}
  }
  \label{tab:processing_speed}
  \vspace{0pt}
\end{table}


By inspection of \tabref{tab:processing_speed} one can see the speed advantage of \ourmethod.
It is important to mention that LSD-SLAM and our system only use a single thread while
ORB-SLAM2 occupies a total of 3 threads. 
All methods are able to process the KITTI datasets in real-time as the stereo camera images are provided at 10Hz.

\section{Conclusion}
\label{sec:conclusion}

In this paper, we presented \ourmethod, a complete stereo visual SLAM system
designed to be easily understood and implemented.
We showed that a proper encapsulation of well known techniques
in separated components with clear interfaces can lead to a very concise SLAM system.
\ourmethod~validates its efficient design by requiring only little computation resources
while maintaining competitive precision and accuracy.
We evaluated our approach on different datasets and provided comparisons to
other existing techniques.

\bibliographystyle{ieeetr}
\bibliography{robots}

\begin{thebibliography}{10}

\bibitem{mur2016orb}
R.~Mur-Artal and J.~D. Tardos, ``{ORB-SLAM2: an Open-Source SLAM System for
  Monocular, Stereo and RGB-D Cameras},'' {\em arXiv preprint
  arXiv:1610.06475}, 2016.

\bibitem{engel2015large}
J.~Engel, J.~St{\"u}ckler, and D.~Cremers, ``{Large-scale direct SLAM with
  stereo cameras},'' in {\em Intelligent Robots and Systems (IROS), 2015
  IEEE/RSJ International Conference on}, pp.~1935--1942, IEEE, 2015.

\bibitem{kummerle2011g}
R.~K{\"u}mmerle, G.~Grisetti, H.~Strasdat, K.~Konolige, and W.~Burgard, ``{g2o:
  A general framework for graph optimization},'' in {\em Robotics and
  Automation (ICRA), 2011 IEEE International Conference on}, pp.~3607--3613,
  IEEE, 2011.

\bibitem{grisetti10titsmag}
G.~Grisetti, R.~K{\"u}mmerle, C.~Stachniss, and W.~Burgard, ``A tutorial on
  graph-based {SLAM},'' {\em Intelligent Transportation Systems Magazine,
  IEEE}, vol.~2, no.~4, pp.~31--43, 2010.

\bibitem{konolige2008frameslam}
K.~Konolige and M.~Agrawal, ``{FrameSLAM: From bundle adjustment to real-time
  visual mapping},'' {\em IEEE Transactions on Robotics}, vol.~24, no.~5,
  pp.~1066--1077, 2008.

\bibitem{pire2015stereo}
T.~Pire, T.~Fischer, J.~Civera, P.~De~Crist{\'o}foris, and J.~J. Berlles,
  ``Stereo parallel tracking and mapping for robot localization,'' in {\em
  Intelligent Robots and Systems (IROS), 2015 IEEE/RSJ International Conference
  on}, pp.~1373--1378, IEEE, 2015.

\bibitem{galvez2012bags}
D.~G{\'a}lvez-L{\'o}pez and J.~D. Tardos, ``Bags of binary words for fast place
  recognition in image sequences,'' {\em IEEE Transactions on Robotics},
  vol.~28, no.~5, pp.~1188--1197, 2012.

\bibitem{hartley2003multiple}
R.~Hartley and A.~Zisserman, {\em {Multiple view geometry in computer vision}}.
\newblock Cambridge university press, 2003.

\bibitem{rosten2006machine}
E.~Rosten and T.~Drummond, ``{Machine learning for high-speed corner
  detection},'' in {\em European conference on computer vision}, pp.~430--443,
  Springer, 2006.

\bibitem{calonder2010brief}
M.~Calonder, V.~Lepetit, C.~Strecha, and P.~Fua, ``{Brief: Binary robust
  independent elementary features},'' in {\em European conference on computer
  vision}, pp.~778--792, Springer, 2010.

\bibitem{smith1986representation}
R.~C. Smith and P.~Cheeseman, ``On the representation and estimation of spatial
  uncertainty,'' {\em The international journal of Robotics Research}, vol.~5,
  no.~4, pp.~56--68, 1986.

\bibitem{schlegel2016visual}
D.~Schlegel and G.~Grisetti, ``Visual localization and loop closing using
  decision trees and binary features,'' in {\em Intelligent Robots and Systems
  (IROS), 2016 IEEE/RSJ International Conference on}, pp.~4616--4623, IEEE,
  2016.

\bibitem{besl1992method}
P.~J. Besl and N.~D. McKay, ``{Method for registration of 3-D shapes},'' in
  {\em Robotics-DL tentative}, pp.~586--606, International Society for Optics
  and Photonics, 1992.

\bibitem{geiger2012we}
A.~Geiger, P.~Lenz, and R.~Urtasun, ``Are we ready for autonomous driving? the
  kitti vision benchmark suite,'' in {\em Computer Vision and Pattern
  Recognition (CVPR), 2012 IEEE Conference on}, pp.~3354--3361, IEEE, 2012.

\bibitem{burri2016euroc}
M.~Burri, J.~Nikolic, P.~Gohl, T.~Schneider, J.~Rehder, S.~Omari, M.~W.
  Achtelik, and R.~Siegwart, ``{The EuRoC micro aerial vehicle datasets},''
  {\em The International Journal of Robotics Research}, vol.~35, no.~10,
  pp.~1157--1163, 2016.

\end{thebibliography}

\end{document}